\documentclass[sigconf]{acmart}
\renewcommand\footnotetextcopyrightpermission[1]{} 
\acmConference{}{}{}
\usepackage[many]{tcolorbox}
\usetikzlibrary{calc}

\usepackage{ graphicx}
\usepackage[ruled,linesnumbered]{algorithm2e}
\usepackage{algorithmic}
\usepackage{paralist}
\usepackage{color}
\usepackage{xcolor}
\usepackage{multirow}
\usepackage{rotating}
\usepackage{hyperref}
\usepackage[mathscr]{eucal} 
\usepackage{subfigure}
\usepackage{booktabs}
\usepackage{xcolor}
\usepackage{tikz}
\usetikzlibrary{snakes}
\usepackage{multirow}
\usepackage{color}
\usepackage{breqn}
\usepackage{balance}
\usepackage{tablefootnote}
\usepackage{empheq} 
\usepackage{moresize}
\usepackage{tabularx}
\usepackage{enumitem}
\setlist{nolistsep}
\usepackage[font=small,labelfont=bf,skip=0pt]{caption}
\usepackage[acronym,nomain]{glossaries}              

\newglossarystyle{notationlong}{%
\setglossarystyle{long}
  {\end{longtable}}%

}

\usepackage{blindtext}

\newglossary[slg]{symbolslist}{syi}{syg}{Symbolslist} 

\makeglossaries                                   
\usepackage{url}

\renewcommand{\algorithmicrequire}{\textbf{Input:}}

\newcounter{ALC@tempcntr}

\renewcommand{\algorithmicrequire}{\textbf{Input:}}

\usepackage{xspace}

\newcommand{\hide}[1]{}

\newcommand{\ben}{\begin{enumerate*}}
\newcommand{\een}{\end{enumerate*}}
\newcommand{\bit}{\begin{itemize*}}
\newcommand{\eit}{\end{itemize*}}

\begin{abstract}
Graphs are one of the most efficacious structures for representing datapoints and their relations, and they have been largely exploited for different applications. Previously, the higher-order relations between the nodes have been modeled by a generalization of graphs known as hypergraphs. In hypergraphs, the edges are defined by a set of nodes i.e., hyperedges
to demonstrate the higher order relationships between the data. However, there is no explicit higher-order generalization for nodes themselves. In this work, we introduce a novel generalization of graphs i.e., K-Nearest Hyperplanes graph (KNH) where the nodes are defined by higher order Euclidean subspaces for multi-view modeling of the nodes. In fact, in KNH, nodes are hyperplanes or more precisely m-flats instead of datapoints. We experimentally evaluate the KNH graph on two multi-aspect datasets for misinformation detection. The experimental results suggest that multi-view modeling of articles using KNH graph outperforms the classic KNN graph in terms of classification performance. 
\end{abstract}
\begin{document}

\title{KNH: Multi-View Modeling with \underline{K}-\underline{N}earest \underline{H}yperplanes Graph for Misinformation Detection}
\author{Sara Abdali}  \email{sabda005@ucr.edu}
\affiliation{%
  \institution{University of California, Riverside}}
\author{Neil Shah} \email{nshah@snap.com}
\affiliation{%
 \institution{Snap Inc.}
 }
\author{Evangelos E. Papalexakis epapalex@cs.ucr.edu}
\affiliation{
  \institution{University of California, Riverside}}
\maketitle
\section{Introduction}

Over the last decades, multiple approaches have been introduced for data representation and classification. Graphs are one of the most efficacious data structures employed extensively by mathematicians and computer scientists for countless applications among which we can mention networks, biological structures, social interaction on social media, recommender systems etc. \cite{biology,social-graph,recommender}. The graph data structures is mostly used to model entities as nodes and their pair wise relationship in form of edges that connect related nodes. Graphs also play a crucial role in machine learning and classification tasks. For instance, we can leverage graphs to model the distance between data points and then predict the class of unknown labeled data based on similarity to other nodes. 
 One of the most effective and widely used graph-based machine learnig modeling approaches is the K-Nearest Neighbours graph (K-NN). In K-NN graph, the entities are  modeled by graph nodes and the K-most similar entities are connected by edges such that the weight of each edge corresponds to the distance or similarity of the connected nodes \cite{Han:2011:DMC:1972541}.
\par The evolution of data storage technologies has made data scientists capable of storing huge volume of information and analyzing the data considering hundreds or thousands of aspects or features. Although accessing more information brings about a more holistic view of data points, the classification task for labeling the datapoints considering different aspects of the data has become a more challenging task. To this end, variety of techniques under the umbrella of ensemble learning approaches have been introduced by machine learning researchers. The ensemble learning approaches aim at combining individual classifiers often designed for one aspect of the data, to create a robuster classifier that merges the decision making process of all individual classifiers\cite{Ensemble_survey,Ensemble_ml}.
\par Unfortunately, the traditional graph structures are not capable enough for multi-view representation of entities. One way that comes into mind is to create multiple of graphs for different aspects and then merge them somehow but considering the high dimensionality of real world datasets, this solution is not only very expensive but also makes the merging step of ensemble learning very complicated. In other words, finding an insightful way to combine all relationships between nodes in terms of thousands of aspects seems almost impossible.  
\par However, an extension of graph models known as hypergraph has been introduced where an edge may connect more than two nodes to illustrate the higher order relationships between the nodes. In fact, instead of a single weighted connection, an edge is  a subset of nodes\cite{hypergraphs1993,Hypergraphs_learning} that are similar in terms of features or distance. There are previous works that leverage hypergraphs for variety of machine learning tasks. For example, in  \cite{hypergraph_application}, for image classification task, Li. et al. propose
an adaptive hypergraph learning method that varies the size of the neighborhood and  generates multiple hyperedges for each sample. Same as previous one, Yu. et al. also leverage  hypergraph data structure for image classification task. In \cite{hypergraghs_multilabel}, Lian. et al. present a hypergraph based formulation for multi-label classification task. In this work, the hypergraph is created to use the correlation information among different labels. There are aslo some previous works that leverage hypergraphs for object detection task \cite{Hypergraph_semi}.
\begin{figure}[!t]
    \begin{center}
    \includegraphics[width = 1\linewidth]{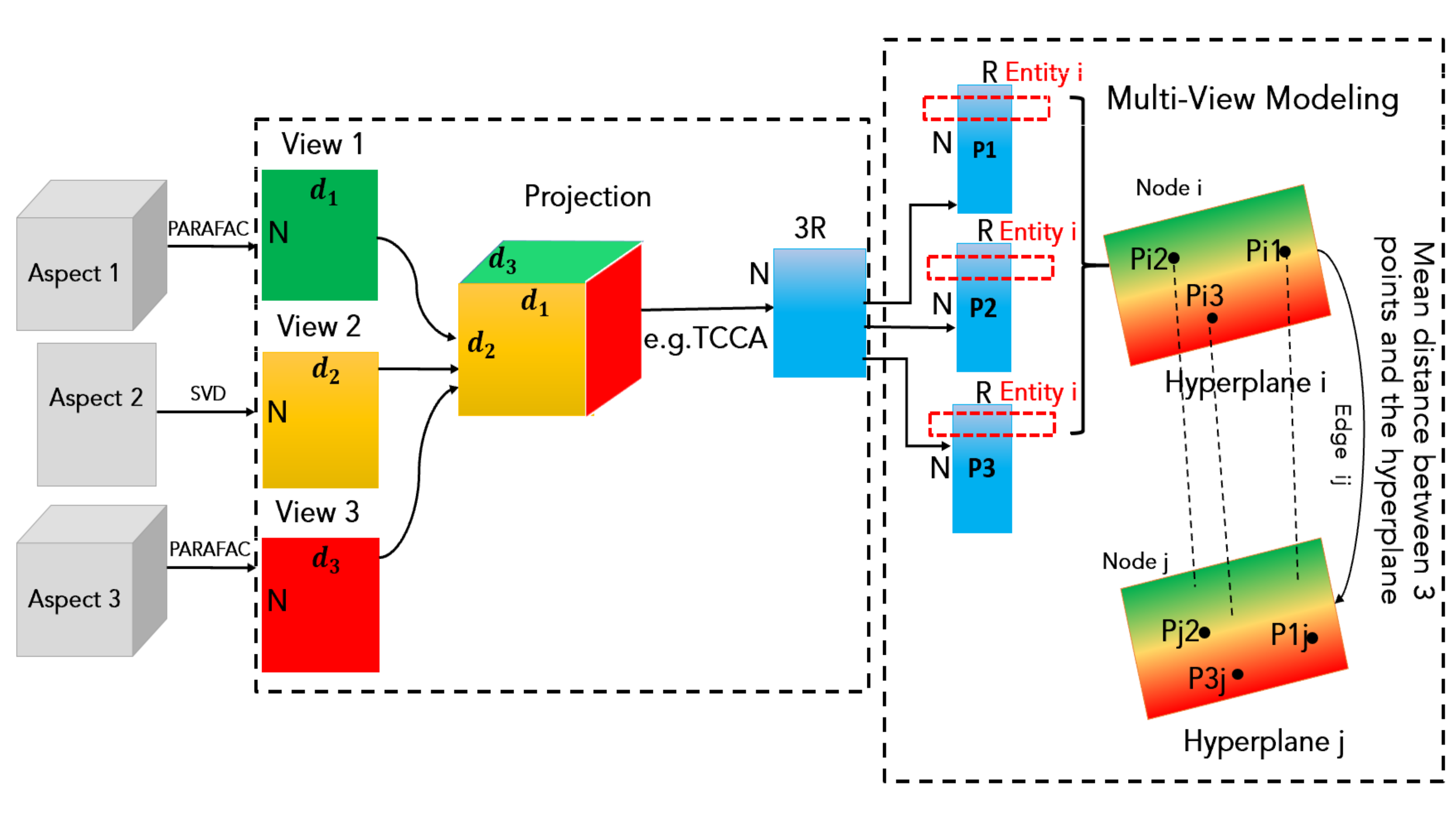}
    \end{center}
    \caption{An overview of proposed  K-NH graph for multi-view modeling and ensemble learning. }
    \label{fig:method}
\end{figure}.
\par Unfortunately, we can not use hypergraphs to illustrate the multi-view representation of nodes. Moreover,  we are not able to define a common "feature space" for entities to predict behaviour of arriving or missing features of data points. The main focus of hypergraphs is on hyperedges or high order relationships between the nodes not multi-view representation of each node which is required for ensemble learning process.
In this work, we propose a novel generalization of learning graphs that aim at defining hypernodes or a common "feature space" for multiple-views of entities  using geometric objects and linear algebra techniques which not only is capable of multi-view modeling of the data but also can be exploited to \textit{predict missing or arriving features. Moreover,  defining a common ensemble feature space for entities enables us to use geometrical-based techniques to calculate intersection, orthogonality, linearity, distance etc. between different objects (entities).} Contrary to hypergraph learning techniques which aim at defining hyperedges and their weights to model high order relationships between nodes, the goal of this work is to model pair-wise relationships but this time by considering multiple views each of which representation of the nodes from different points of view. In other word, we want to introduce a generalization of K-NN graphs which can be used for ensemble learning tasks.  
\par To this end, we propose to first capture the entity representations in different feature spaces, henceforth refer to as views and then we map these views to a new shared space so that we can use the mapped views for defining higher order geometric objects which represent a holistic view of entities.
\par The contribution of this work are as follows:
\begin{itemize}
    \item \textbf{A novel graph based modeling for multi-view representation of data points using geometric objects} In this work we introduce a generalization of K-NN graphs where contrary to traditional K-NN graphs that each node is a datapoint in $N$-dimensional space, there are hypernodes defined by $M$-dimensional flats $M < N$ (subspaces in $N$-dimensional space). This hypernode (subspace) are defined by multiple datapoints (views) for each node and present a holistic view of each entity that can be used for prediction of missing or arriving features.
    \item \textbf{A novel decomposition-based pipeline for ensemble learning}
    We introduce a novel decomposition-based pipeline that leverages K-NH graphs of this work for ensemble learning and multi-view classification of data.This pipeline consists of decomposition (CP or SVD), tensor canonical analysis (TCCA), graph modeling and distance calculation.
    \item \textbf{Experimenting on real world problems and applications} In this work, we examine the K-NH modeling and classification pipeline for classification of two real world datasets including textual, user and social context views.
\end{itemize} 
\par The organization of the paper is as follows:
\par We first present the related work in section \ref{sec:Related work} and then we discuss the mathematical background required for K-NH modeling and the classification pipeline. Next, we state the problem formulation and then in section \ref{sec:method} we describe the proposed K-NH modeling and classification pipeline. In Section \ref{sec:experiment}, we examine the proposed method on two real world datasets and finally we conclude.
\section{Related work}

\subsection{Ensemble Learning for Fake News Detection}
The majority of misinformation detection approaches focus on a single aspect of the data and mostly the article content \cite{shu2017fake,wu2017gleaning}. There are also works that leverage other aspects like user features \cite{wu2018tracing}, and temporal properties \cite{kumar2018false}. However, there exist few ensemble approaches that consider all different aspects simultaneously. For Instance, in \cite{Beyond} the authors propose an ensemble model by merging a bag of words embedding, user-user, user-article and publisher-article interactions. In another work \cite{dEFEND}, news contents and user comments are consolidated to detect the misinformation jointly. Another example is \cite{HiJoD}, where content-based, social-context in form of hashtags and website features are leveraged to create manifold patterns for multi-aspect detection of misinformation. In this work, we leverage the promising aspects introduced in both \cite{Beyond} and \cite{HiJoD} but this time with a different and novel multi-aspect modeling and formulation.
\label{sec:Related work}
\subsection{Hypergraph Learning}
The hypergraphs are one extension of graph models in which  an edge can connect more than two nodes. In other words, an edge is defined as a subset of nodes\cite{hypergraphs1993,Hypergraphs_learning} that share same (similar) feature.
In contrast to traditional graph-based learning methods which only model the pairwise relationship
between entities, the hypergraph leverage hyperedges to model higher-order relationships between the entities. In previous work, the hypergraph learning has been used for variety of machine learning applications. For instance, in  \cite{hypergraph_application} Li. et al. propose to model an image as a hypergraph that leverages  hyperedges to capture the contextual features of the pixels. In another work, Lian. et al. construct a hypergraph to exploit
the correlation information among labels for multi label classification task\cite{hypergraghs_multilabel}. In
\cite{Hypergraph_semi} Yu. et al.  propose an adaptive hypergraph based method for classification of images. Moreover, there are previous works that leverage hypergraphs for object detection tasks\cite{object_detection2,objectdetection}.
In hypergraphs, the main focus is to define hyperedges and the weights to model high order relationships. Although there are some unsupervised work using affinities within the hyperedges, \cite{unsupervised_hypergraph}, hypergraphs, do not have exploratory capabilities to define a common "feature space" to predict behaviour of arriving data points (nodes) or predicting the missing data points using this common space. Moreover, finding the weights for the hyperedges is a challenging task and requires complicated optimization and regularization techniques like graph Laplacian, $\mathbf{L}_1$ and $\mathbf{L}_2$ regularizers
 \cite{Hypergraph_semi,l1_hypergraph}. In this work, we try to present a generalization of learning graphs that mostly focuses on defining hypernodes or a common "feature space" for multiple-views of entities in dataset which not only is capable of multi-view modeling of the data but also can be exploited to predict missing or arriving data.
 \section{Background}
\label{sec:method}
In this section, we first present mathematical background requires for the proposed method and then we discuss the problem definition and proposed K-Nearest Hyperplanes Graph (KNH).
\begin{table}[th]
\centering
\begin{tabular}{
||p{2cm}p{4.5cm}||}
 \multicolumn{2}{c}{\textbf{Table of Notations}} \\
 \hline
 \textbf{Symbol} &\textbf{Definition}\\
 \toprule
 \hline
 \textbf{$\mathcal{X}$},\textbf{X},x& Tensor,Matrix,vector\\
 $\circ$&Outer product\\
 $\times$&Cross product\\
$Cov(x,y)$&Covariance x and y\\
$E(x)$&Mean x\\
$\rho=Corr(x,y)$&Correlation between x\&y \\
$C_{xx}$&Variance matrix of vector x\\
$C_{xy}$&Covariance matrix of vectors x\&y\\
$C_{1 2 \cdots m}$&Covariance Tensor\\
$h_{x}$&Canonical vector\\
$z_{x}$&Canonical variable\\
\bottomrule
\end{tabular}
\caption{Symbols and Definitions}
\label{table:notation}
\end{table}
\vspace{1mm}
\subsection{Matrix and Tensor Decompositions}
\par A tensor is an array with three or more than three dimensions where the dimensions are usually referred to as modes\cite{Papalexakis:2016,Tensor}. In linear algebra, there is a factorization algorithm known as 
Singular Value Decomposition (SVD) in which we can  factorize a matrix $\boldsymbol{X}$ into the
product of three matrices as follows:
\begin{dmath}
\boldsymbol{X} \approx\mathbf{U} \boldsymbol{\Sigma} \mathbf{V}^{T}   
\end{dmath}
\par where the columns of U and V are orthonormal and
the matrix $\boldsymbol{\Sigma}$ is a diagonal with positive real entries.
Using rank $\mathbf{R}$, SVD decomposition we can represent a matrix as a summation of $\mathbf{R}$
rank 1 matrices as follows:
\begin{dmath}
\boldsymbol{X}
\approx \Sigma_{r=1}^{R} \sigma_r \mathbf{u}_r \circ \mathbf{v}_r 
\end{dmath}
\par The Canonical Polyadic (CP) or PARAFAC decomposition is an extension of SVD for higher mode matrices i.e., tensors \cite{harshman1970fpp}. Indeed, CP/PARAFAC factorizes a tensor into a summation of rank-one tensors. For instance, a three-mode tensor is decomposed into a sum of outer products of three vectors as follows:
\begin{dmath}
 \textbf{$\mathcal{X}$} \approx \Sigma_{r=1}^{R} \mathbf{a}_r \circ \mathbf{b}_r \circ \mathbf{c}_r
\end{dmath}
where $\mathbf{a}_r \in \mathbb{R}^{I}$, $\mathbf{b}_r \in \mathbb{R}^{J}$, $\mathbf{c}_r \in \mathbb{R}^{K}$ and the outer product is given by \cite{Papalexakis:2016,Tensor}:  
\begin{dmath}
(\mathbf{a}_r , \mathbf{b}_r , \mathbf{c}_r)(i,j,k)=\mathbf{a}_r(i)\hspace{1mm} \mathbf{b}_r(j)\hspace{1mm} \mathbf{c}_r(k) \hspace{1mm} \forall{i,j,k}
\end{dmath}
\vspace{2mm}
Factor matrices are defined as $\mathbf{A} = [\mathbf{a}_1 ~ \mathbf{a}_2 \ldots \mathbf{a}_R]$, $\mathbf{B} = [\mathbf{b}_1 ~ \mathbf{b}_2 \ldots \mathbf{b}_R]$, and $\mathbf{C} = [\mathbf{c}_1 ~ \mathbf{c}_2 \ldots \mathbf{c}_R]$
 where $R$ is the rank of decomposition or the number of columns in the factor matrices.The optimization problem for finding factor matrices is as follows:
 
\begin{dmath}
 \min_{A,B,C} ={{\lVert}  \textbf{$\mathcal{X}$} - \Sigma_{r=1}^{R} \mathbf{a}_r \circ \mathbf{b}_r \circ \mathbf{c}_r{\rVert}}^2
\end{dmath}
\par One effective way for solving the optimization problem above is to use Alternating Least Squares (ALS) which solves for any of the factor matrices by fixing the others  \cite{Papalexakis:2016,Tensor}.
\subsection{Canonical Correlation Analysis (CCA)}
In 2-dimensional space the correlation between two vectors $x,y$ is defined as follows \cite{CCA,TCCA}:\\
\begin{dmath}
 \rho=Corr(x,y)=\frac{cov(x,y)}{\sigma_x\sigma_y}
\end{dmath}
Since $Cov(x,y)=E(xy)-E(x)E(y)=E(xy)$, if we suppose the vectors are centered around the mean, then $E(x)$ and $E(y)$ are zero and $\rho$ is going to be\cite{TCCA}:
\begin{dmath}
  \rho=\frac{E(xy)}{\sqrt{E(x^2)E(y^2)}}
\end{dmath}
\par There is a technique known as Canonical Correlation Analysis or CCA which we can use to find canonical vectors $h_x, h_y$ such that if we project two vectors $x$ and $y$ using these two canonical vectors into canonical variables $z_x,z_y$, the correlation between $z_x$ and $z_y$ is maximized \cite{CCA,TCCA}:
\begin{dmath}
argmax \rho_{z_1,z_2}=corr(z_1,z_2)
  =\frac{E(h_x^Txy^Th_y)}{\sqrt{E(h_x^Txx^Th_x) E(h_y^Tyy^Th_y)}}
   = \frac{h_x^TC_{xy}h_y}{\sqrt{h_x^TC_{xx}^Th_x h_y^TC_{yy}h_y}}
   \end{dmath}
Where $C_{xx}=XX^T$,$C_{yy}=YY^T$ are variance matrices and $C_{xy}=XY^T$ is covariance matrix of vectors $x$ and $y$.
\subsubsection{Tensor Canonical Correlation Analysis (TCCA)}
When we have more than two variables, we can also define the optimization problem above as a minimization problem where we aim at minimizing the pairwise distance between the variables. So, the generalized form of the CCA can be redefined as follows \cite{TCCA}:
\begin{dmath}
argmin_{h_p} \frac{1}{2m(m-1)}\Sigma^{m}_{p,q=1} \| X_p^Th_p-X_q^Th_q\|^2
\label{equ:TCCA_opt}
\end{dmath}
\par As we know $C_{xy}=XY^T$ is equal to covariance matrix of $x$ and $y$. In higher dimensional space we can also define variance matrix $C_{pp}$ and covariance tensor $C_{1\cdots m}$ as follows \cite{TCCA}:
\begin{dmath}
C_{1 2 \cdots m}= \frac{1}{M}\Sigma^{m}_{n=1} x_{1n}\circ x_{2n}\circ \cdots \circ x_{mn}
\end{dmath}
\begin{equation}
C_{pp}= \frac{1}{M}\Sigma^{m}_{n=1} x_{pn}x_{pn}^T \hspace{2mm} p \in {1\dots m}
\end{equation}
\par We can show that higher order canonical correlation can be computed by CP/ALS optimization probelm. For proof you can refer to \cite{TCCA}.
\subsection{Hyperplanes and flats in n-dimensional space}
A hyperplane in an n-dimensional space $V$ is an $n-1$ dimensional subspace which is defined by following linear equation \cite{Topological}:
\begin{dmath}
a_1(x_1-{x_1}^')+a_2(x_2-{x_2}^')+\dots+a_n(x_n-{x_n}^')=0
\end{dmath}
\par Where the vector $(a_1,a_2,\dots,a_n)$ is a normal vector perpendicular to the hyperplane and $({x_1}^',{x_2}^',\dots,{x_n}^')$ is a point on the hyperplane. Therefore, we can rewrite the linear equation  of hyperplane as\cite{Topological}: 
\begin{dmath} 
a_1x_1+a_2x_2+\dots+a_nx_n=d
\label{equ:plane}
\end{dmath}
\par Given $n$ datapoints we can uniquely define a hyperplane in an n-dimensional space. The distance from a point $({x_1}^',{x_2}^',\dots,{x_n}^')$ to a hyperplane is defined as follows\cite{Topological}:
\begin{dmath}
d_{point-hyperplane}=\frac{| a_1{x_1}^'+a_2{x_2}^'+\dots+a_n{x_n}^'+d|}{\sqrt{{a_1}^2+{a_2}^2+\dots+{a_n}^2}}
\label{equ:pointplane_dis}
\end{dmath}
A flat or Euclidean subspace is any lower dimension subspace in that space. For instance, flats in 4-dimensional space are points, lines, and planes.
We can described a flat in n-dimensional space by a system of linear parametric equations. For example, the equation of a line in n-dimensional space is equal to:
\begin{equation}
x_1=a_{1}t+b_1, x_2=a_{2}t+b_2,\dots, x_n=a_{n}t+b_n
\label{equ:line_flat}
\end{equation}
\par Then we can use the Euclidean distance to calculate the distance from a point to a 2-flat (line). For instance, we can calculate the distance from  point $P_0$ to a 2-flat (line) defined by 2 points $P_1$ and $P_2$ in 3-dimensional space as:
\begin{dmath}
\label{equ:point_line_distance}
  d_{point-line}=\frac{|(p_2-p_0) \times (p_1-p_0)|}{|p_2-p_1|}
\end{dmath}
\par Where $\times$ is the cross product of two vectors $(p_1-p_0)$ and $(p_2-p_0)$.
\par Just like the previous one, we can define a 3-flat (plane) in n-dimensional space as follows:
\begin{equation}
x_1=a_{1}t_1+b_1t_2+c_1, x_2=a_{2}t_1+b_2t_2+c_2,\dots, x_n=a_{n}t_1+b_nt_2+c_n
\end{equation}
\subsection{$k$-nearest-neighbor graph}
We can model entities in a dataset using a $k$-nearest-neighbor graph in which each entity is a node or a datapoint in feature space and the edges between the datapoints represent the distances or similarities between the entities. \cite{Han:2011:DMC:1972541}. 
\par What if there are multiple of views for each entity in the dataset, each of which represents the entity with respect to a specific aspect of it? The idea of this work is to find a holistic representation that comprises all different views of the entity and finding a way for calculating the  distances between these manifold representation of each entity. In next section we define a novel approach for generalizing the KNN graphs and a new way for measuring the similarity between the entities.

\section{Problem Formulation}
\label{sec:problem}
The problem formulation of multi view modeling and classification using K-Nearest Hyperplanes graph is as follows:
\vspace{1mm}
\tcbset{colback=orange!4!white,colframe=orange!30!white}
\begin{tcolorbox}[width=1\linewidth]
\textbf{Given} a dataset comprising $N$ entities and $M$ matrices of size $N\times d_m$,$m=1,\dots,M$ for $M$ views of the data such that row $i$ of matrix $m$ corresponds to a $d_m$-dimensional representation of entity $i$ with respect to view $m$.\\ \textbf{Find} a representation for the entities 
\\ \textbf{Such that}  the manifold structures are preserved when used for modeling and classification of those entities.
\end{tcolorbox}
One simple solution that comes into mind is to stack views into a long vector and use KNN graph for modeling and classification. But by doing so, we may destroy potentially useful structures.
We address this problem by defining a $M$ dimensional flat in $R$ dimensional space for each entity where $R$ is the dimensionality of view matrices after projecting into a common space. For example, if we have 2 view matrices we can model each entity by a line and if we have 3 views, we model entities with a plane. These flats are generalized form of points (nodes) in KNN graph. We can then leverage geometrical properties of hyperplanes to calculate a manifold distance between entities which can be shown as graph edges. 
As we will see in upcoming experiments, retaining the proposed representation results in better quality in downstream classification tasks.
The details are described in next section.
\section{Hyperplane modeling and K-nearest hyperplanes graph}
In what follows the hyperplane modeling and classification will be described step by step.
\subsection{Modeling the Aspects using Tensor/Matrix and Decomposing Aspects into View Matrices}
Matrices and tensors are common tools for modeling entities in feature space. For instance, using the well known bag of word matrix we model documents in word space. Likewise, for multi aspect modeling of the entities, we leverage tensors such that one mode of the tensor correspond to entities and other modes represent different aspects that the entities are defined by. To capture the hidden patterns of the entities with respect to the considered aspect(s), we decompose the matrix(tensor) into factor matrices as described in previous section. Having this in mind, the very first step of the proposed approach is to decompose $M$ models of the entities (matrices or tensors) into $M$  entity mode factor matrices henceforth referred to as view matrices each of which of size $N \times d_m$ where $N$ is the number of entities and $d_m$ is the size of latent pattern space defined by rank of decomposition. In fact, each view comprises latent patterns of the entities with respect to the considered aspect.
\subsection{Projecting Views into a Common Space}
Previous step provides us with $M$ pattern matrices of size $N \times d_m$, $m=1,\dots,M$ for $M$ different views of the data. Now, we want to leverage all these view matrices to create a manifold description of entities. In fact, the goal is to define a new space that consolidates all $M$ representation of the entities. Since these matrices represent the entities in different spaces, we need to find a way to project all theses different representations into a common space such that the correlation between all representations is maximized. One solution that comes into mind for this requirement is the Canonical Correlation Analysis or CCA as discussed earlier. Likewise, if we have more than 2 vectors for each entities corresponding to more than 2 view matrices, we can leverage TCCA  or higher order CCA to maximize the correlation between the rows of $M$ views. To this end, we first create a tensor $\mathcal{X}$ of size $d_1\times d_2 \dots \times d_m$ out of all $M$ matrices which is equivalent to the covariance tensor. Then we leverage TCCA algorithm as explained earlier, to project all views into a new space. The rank of decomposition is equal to the dimension of the new space. As an example, suppose we have three view matrices. We define a 3-mode covariance tensor as follows:
\begin{dmath}
\boldsymbol{\mathcal{C_{123}}} \approx \Sigma_{r=1}^{R} \mathbf{m_1}_r \circ \mathbf{m_2}_r \circ \mathbf{m_3}_r
\end{dmath}
 Where $C_{123}$  is the covariance tensor and $m_1$ to $m_3$ are view matrices and $R$ is the rank of decomposition for finding the maximally correlated variables. Now, we leverage TCCA algorithm to solve the equation \ref{equ:TCCA_opt}.
\subsection{Creating K-Nearest Hyperplane Graph for Classification}
Previous step results in $N$, $M$-flats
in $R$-dimensional space, each of which a manifold representation of entity $i$, $i={1,\dots,N}$. We can create a generalized K-NN graph in such a way that each node of the graph is a $M$-flat in $R$-dimensional space and the edges between the nodes show the multilateral similarity between the flats (nodes).
The question that raises here is: "how to calculate the distances between the hyperplanes?" because if we are to use the Euclidean distance between the hyperplanes, they should be parallel, otherwise the distance between them is equal to zero.
\par
\begin{figure}[H]
    \begin{center}
    \subfigure[Similarity based on angles between normal vectors. As depicted, angle between the hyperplanes i.e., angle between the normal vectors is not a proper metric to measure the similarity e.g., $P_i$ is closer to $P_j$ than $P_k$, but the angle both plane form with plane $i$ is equal.]{\includegraphics[width = 0.4\linewidth]{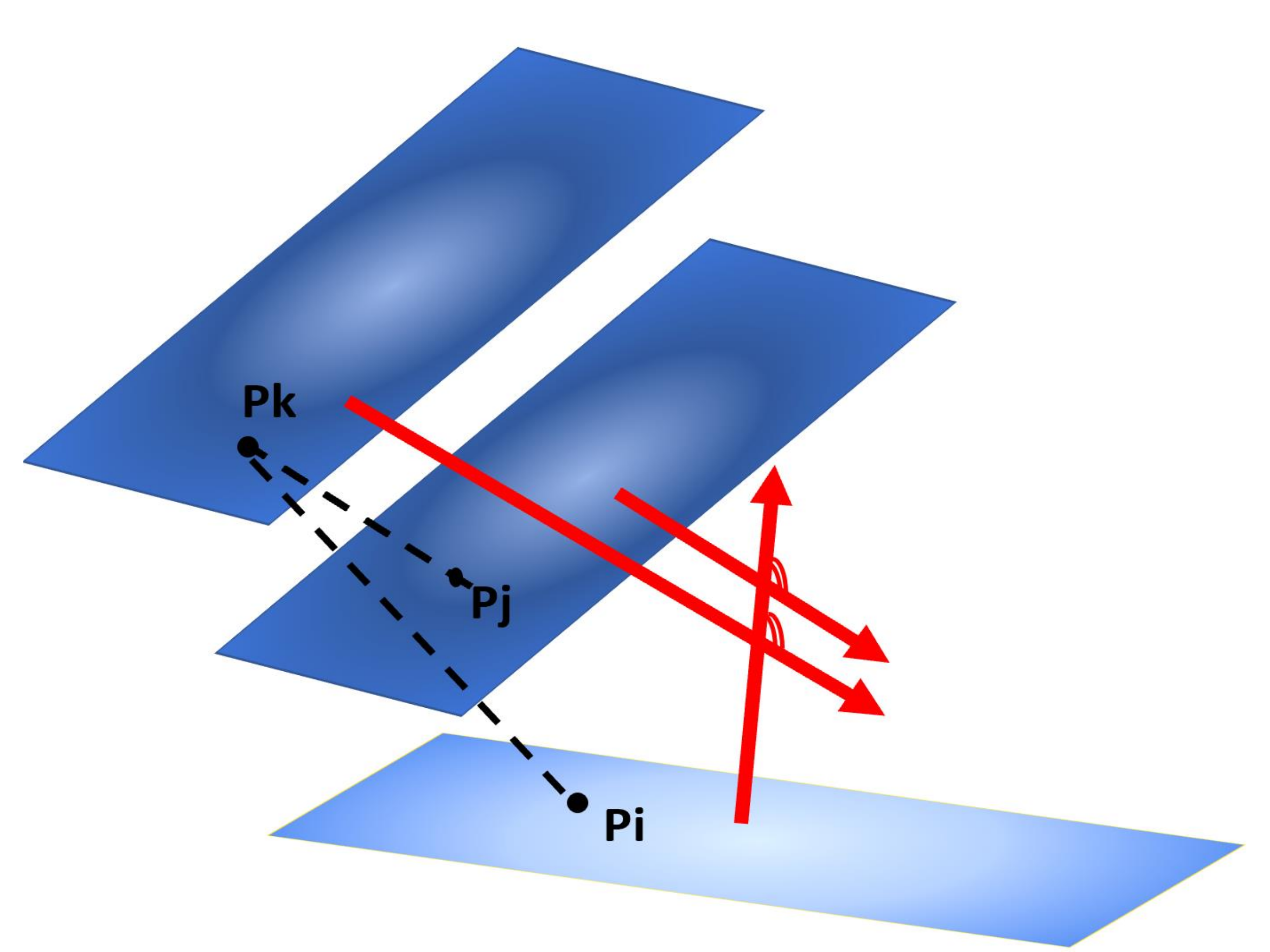}}
    \hspace{6mm}
    \subfigure[Similarity based on point-plane Euclidean distance of points of one plane to another plane. The closer the points are to the intersection of the hyperplanes, the smaller the distance gets.]{\includegraphics[width = 0.4\linewidth]{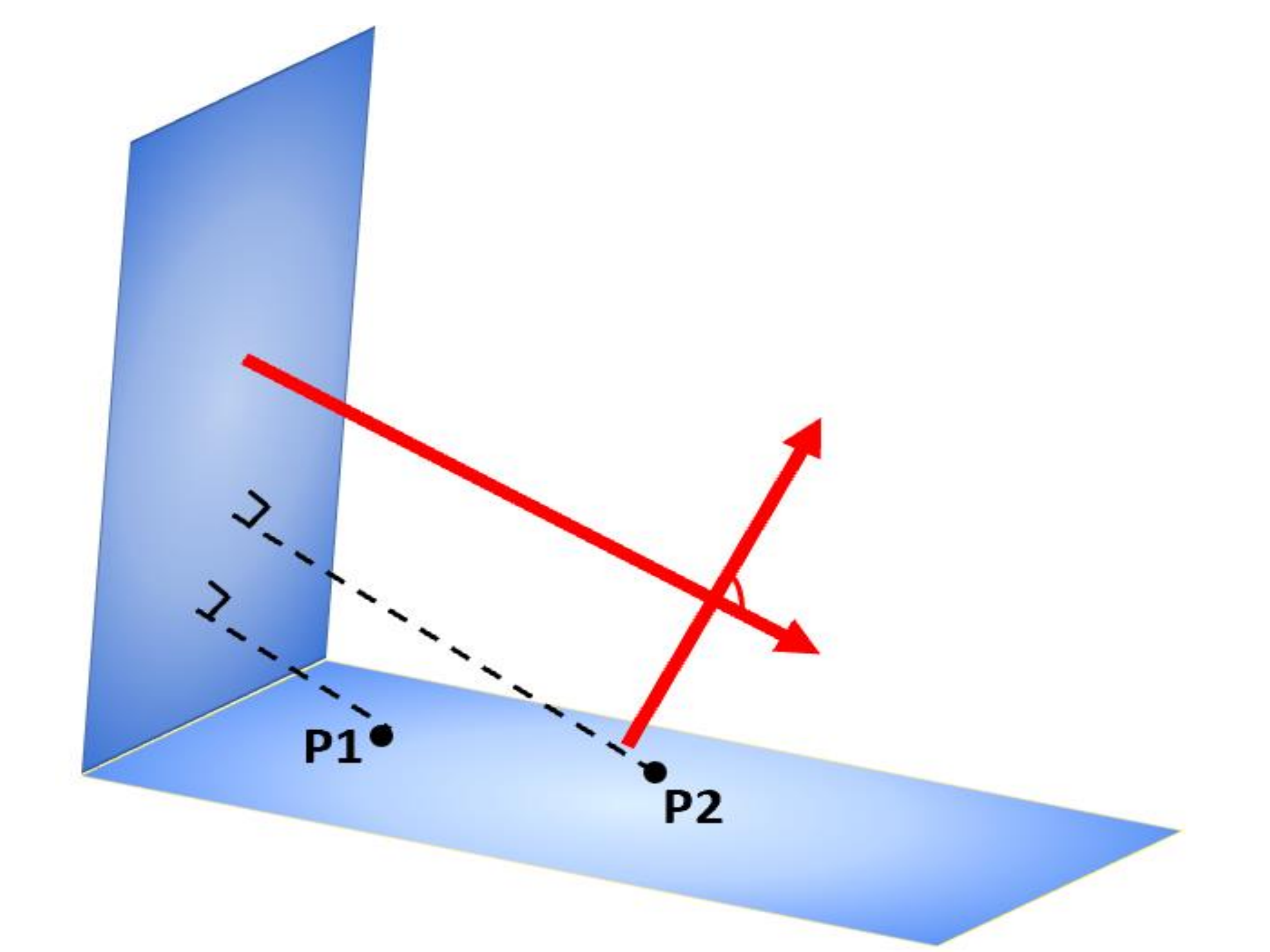}}
    \end{center}
    \caption{Comparing different approaches for measuring similarity between hyperplanes.}
    \label{fig:distance}
\end{figure}

 One way that comes into mind is to calculate the angle between the hyperplanes which is equal to the angle between the normal vectors of the hyperplanes, but lets consider the situation demonstrated in Figure. \ref{fig:distance} part a, where plane $j$ and $k$ are parallel to plane $i$ so, they form the same angle with plane $i$. In this situation, there might be a point $P_i$ lying on plane $i$ which is closer to a point $P_j$ on plane $j$ than a point $P_k$ on plane $k$. Thus, the angle scenario is not capable to capture this difference. But if we consider the point-hyperplane distance using \ref{equ:pointplane_dis}, then we are able to capture an insightful difference illustrated in  \ref{fig:distance} part b. The closer the points are to the \emph{intersection of the hyperplanes}, the smaller the $d_{point-hyperplane}$ gets.
 .
\par Having justification above in mind, we define the following distance as the distances between the hyperplanes in KNH graph. we use the mean of Euclidean distances between each of $M$ datapoints defining hyperplane $i$ and hyperplane $j$ as mentioned in equation \ref{equ:pointplane_dis}. For instance, in $R$-dimensional space we can define the weight $d_{ij}$ of the edges  using Algorithm. \ref{alg:KNH}.
\par In Figure \ref{fig:method} and algorithm \ref{alg:KNH} an overview of the KNH approach is demonstrated.
\begin{algorithm}[t]
\SetAlgoLined
\algorithmicrequire{$2$ embedding of size $N \times d_m$ }
\par \KwResult{ a $K$-nearest hyperplane graph}
$\slash \slash$ Projecting all embeddings into a common R-dimensional space where $w$ is a $N \times 2R$ matrix\\
$W=CCA(M_1,M_2,R)$\\
$P1=w(:,1:R)$\\
$P2=w(:,R+1:2R)$\\
$\slash \slash$ Defining the Line\\
$x_1=a_{1}t+b_1, x_2=a_{2}t+b_2,\dots, x_n=a_{R}t+b_R$\\
\For{all $i, i={1,\dots,N}$}{
\For{all $j, j={i,\dots,N}$}{
   $p_{1j}= P_1(j,:) - P_1(i,:)$\\
   $p_{2j}= P_2(j,:) - P_1(i,:)$\\
   $p_{12i} = P_2(i,:) - P_1(i,:)$\\
$t_1 = dot( p_{1j}, p_{12i})/dot( p_{12i}, p_{12i})$\\
$t_2 = dot( p_{2j}, p_{12i})/dot( p_{12i},p_{12i})$\\
   $d_1 = ( p_{1j}- t_1 * p_{12i})$\\
   $d_2 = (p_{2j} - t_2 * p_{12i})$\\
   $d_1= sqrt(sum(d_1.^2))$\\
   $d_2=sqrt(sum(d_2.^2))$\\
   $d(i,j)=(d_1+d_2)/2$\\
}
}
Create_Graph$(d)$
\caption{K-Nearest Hyperplanes modeling using 2-views in R-dimensional space}
\label{alg:KNH}
\end{algorithm}
 
\subsection{Complexity Analysis}
The time complexity of KNH method depends on the time complexity of the TCCA and the construction of the graphs which consists of construction of the nodes and calculating the weight of the edges. As discussed in \cite{TCCA}, the time complexity of TCCA is independent of the number of instances and can be scaled for large size problems and the space and time complexity of the approach are $O(N^m)$ and $O(tr N^m)$ respectively\cite{TCCA}. To define the nodes, we need to calculate the normal vectors which is equivalent to  calculating $N$ cross product each of which of size $M$ or number of views $O(NM)$. The complexity of calculating the edges is same as the time complexity of KNN classification and is $O(N^2)$. 
\section{Experiments}
\label{sec:experiment}
In this section, we empirically evaluate the eﬀectiveness of the proposed KNH method against traditional KNN for multi-view modeling and classification task. We experiment on a 2-aspects document-publisher dataset extracted from Twitter's tweets \footnote{https://github.com/Saraabdali/Fake-News-Detection-_ASONAM-2018} and another 2-aspects news article dataset extracted from FakeNewsNet dataset\footnote{https://github.com/KaiDMML/FakeNewsNet} but this time we experiment on different sets of features, i.e., user-news interaction aspects and the publisher-news interaction aspect. Henceforth, we refer to the first dataset as Twitter dataset and to the second dataset as politifact dataset. We first, introduce the details of each dataset and the extracted aspects and then we present the experimental results.

\subsection{Implementation}
We implemented both experiments described above in Matlab using Tensor Toolbox version 2.6 \cite{matlab}. For rank of decomposition (dimensionality) $r_m$ of each view and the number of nearest neighbors $K$ we grid searched the values between range 1-50 for $r_m$ and $1-30$ for $K$. Later on, we will show the classification trends for different ranks and number of neighbors. We measured the effectiveness of all methods using  average precision, recall, F1 score and accuracy for 10 runs of each method.
\begin{figure*}[!th]
\centering
\subfigure[Average F1-score for 10 runs of decomposition using $k$=15 when modeling the articles by KNN and KNH graphs. The results suggest that KNH leads to higher performance especially when we increase the rank.]{\includegraphics[width = 0.40\linewidth]{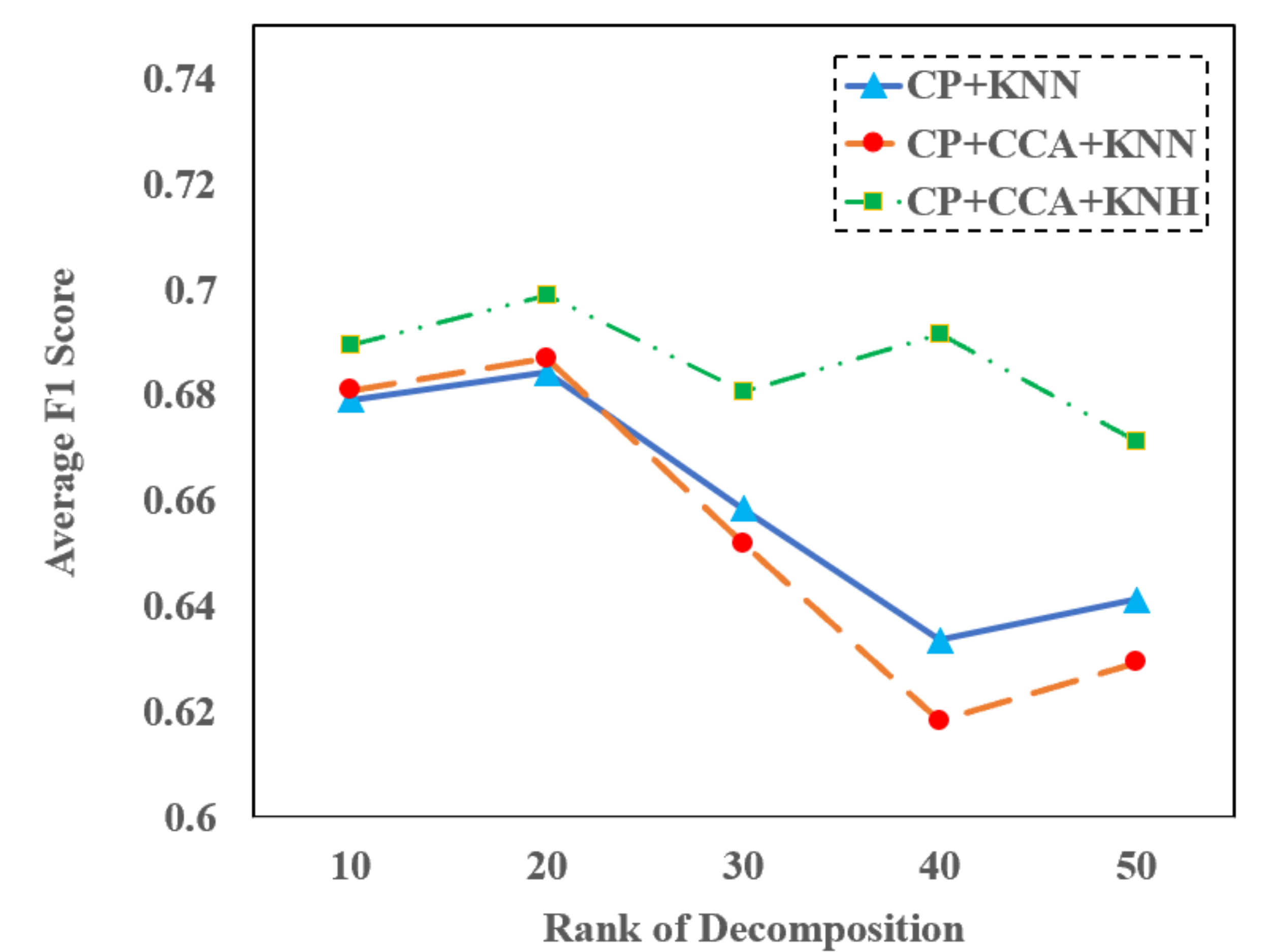}}
\hspace{15mm}
\subfigure[Average F1-score for 10 runs of decomposition using $R$=20 when modeling the articles by KNN and KNH graphs. As depicted, for all number of neighbors KNH results in higher classification performance.]{\includegraphics[width = 0.40\linewidth] {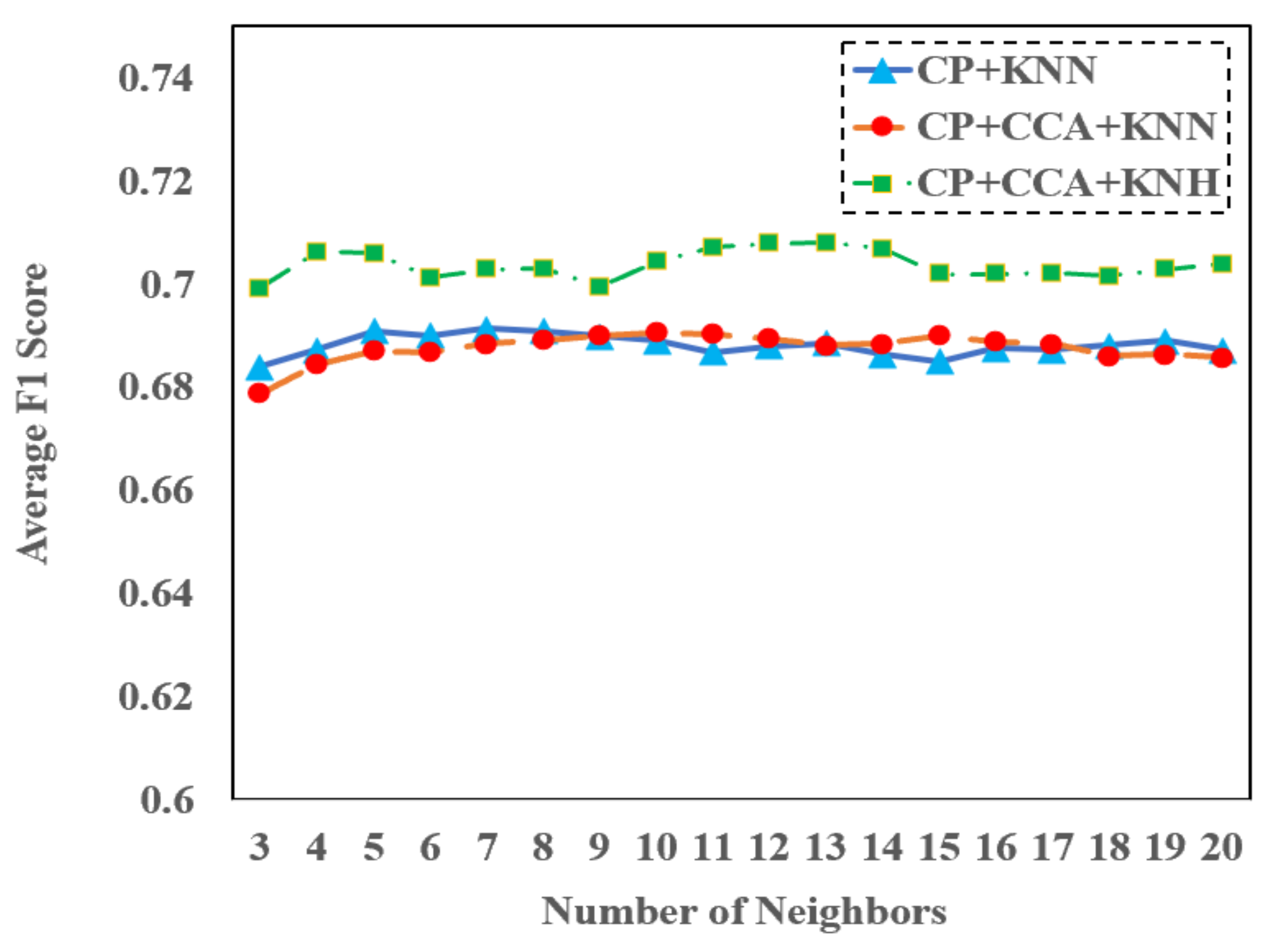}}
\caption{Average F1 score of KNH and KNN modeling for different ranks and number of neighbors.}
\label{fig:twitter}
\end{figure*}
 \begin{figure}[h]
    \begin{center}
    \includegraphics[width = 0.8\linewidth]{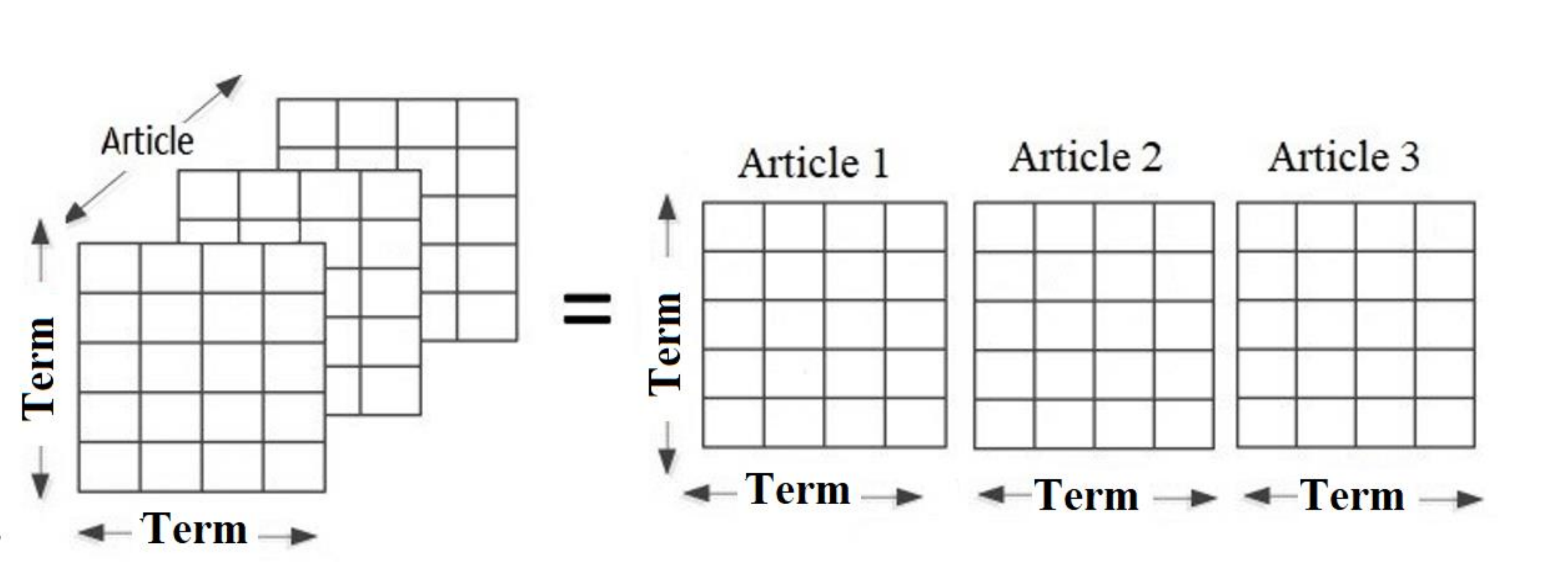}
    \end{center}
    \caption{(Term, Term, Article) tensor to model the textual patterns form by co-occurrence of the words.}
    \label{fig:TTA}
\end{figure}
 \begin{figure}[h]
    \begin{center}
    \includegraphics[width = 1\linewidth]{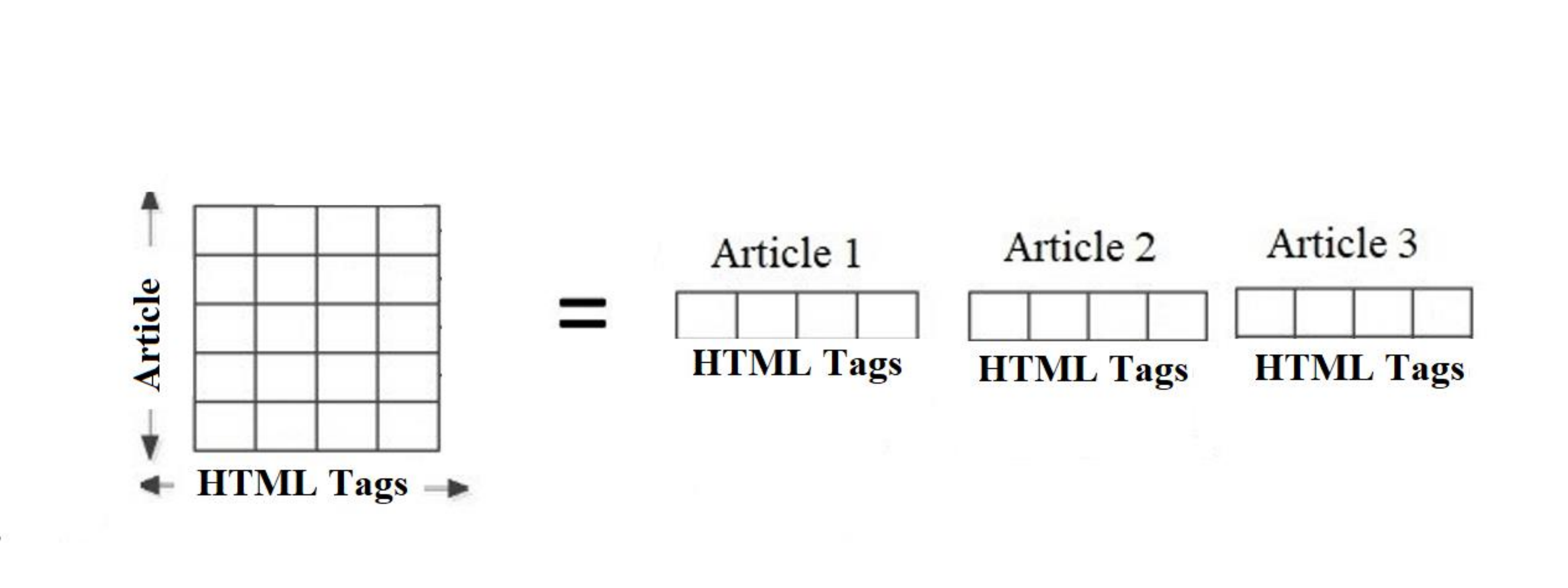}
    \end{center}
    \caption{(Article, Domain feature) Matrix to model the publisher domain features.}
    \label{fig:TAGS}
\end{figure}
\subsection{Experiment 1: Article Classification using 2-Flats (Lines) Created by Textual Content and Domain Aspects}
\subsubsection{Description of dataset and aspects}
As mentioned earlier, for the first experiment, we use the dataset introduced in \cite{ASONAM2018,HiJoD}. This dataset comprises multi-aspect information about news articles and the Twitter tweets shared these articles as URL links. In this dataset, the labels are extracted using the BSDetector Google Chrome extension \footnote{http://bsdetector.tech/} which is a {\em crowd-sourced} toolbox. In aformentioned works, the bias, clickbait, conspiracy, fake, hate, junk science, rumor, satire, and unreliable categories as considered as misinformative articles. In this work, we also follow the same strategy. Moreover, to prevent the domain bias discussed in \cite{HiJoD}, as suggested, we select one article per domain. Thus, we created a relatively balanced sample by randomly selecting one articles per domain as described in Table. \ref{table:dataset1}.

\begin{table}[H]
\footnotesize
\centering
\begin{tabular}{
||p{4cm} p{3cm}||}
 \hline
 \multicolumn{2}{||c||}{\textbf{Twitter dataset}} \\
 \toprule
 \hline
 \textbf{Features}&\textbf{Total Number}\\
 \hline
 words&18853\\
 Domains &652\\ 
 Article&335 (Real)/317 (Fake)\\
\bottomrule
\end{tabular}
\caption{Twitter dataset description}
\label{table:dataset1}
\end{table} 

As the base case i.e., 2-view classification of articles which corresponds to 2-flat (line) modeling, we leverage the most promising aspect models i.e., TTA and Tags introduced in \cite{HiJoD}. The description of the models is as follows:
\begin{itemize}
\item \textbf{(Term, Term, Article) Tensor}:
As suggested in \cite{Hosseinimotlagh2017UnsupervisedCI,ASONAM2018} different classes of news articles, i.e., misinformative and real classes tend to have some common words that co-occur within the text. The co-occurrence of the words forms some patterns which is shared between different categories of the articles. Thus, we use a tensor proposed by \cite{Hosseinimotlagh2017UnsupervisedCI,ASONAM2018} to model co-occurrence of the article words. In this model, we find the co-occurred words by sliding a window across the article text. This yields to a word by word matrix for each article. By stacking all these matrices, we create a three mode tensor where the first and the second modes correspond to the words and the third mode corresponds to the articles as illustrated in Figure \ref{fig:TTA}. We use this model because as shown in \cite{HiJoD} it outperforms some state-of-the-art text based modeling in terms of classification performance and could be applied to many document and text classification tasks. 
\item \textbf{(Article, Domain feature) Matrix}: Another existing information in this dataset is the publisher web features in form of HTML tags. The rationale behind using these features is that different domains have different web styles. For instance trustworthy publishers like BBC and CNN tend to have standard webpages while unreliable resources often have messy webpages full of Ads, pop-ups etc. In \cite{HiJoD}, it has been shown that taking into account this information leads to a very promising classification performance. Therefore, We created a matrix out of the HTML features of the domains as demonstrated in Figure\ref{fig:TAGS}.
\end{itemize}

Henceforth, we refer to the word and publisher tensor as $\mathcal{X}_{TTA}$ and $\textbf{X}_{TAGS}$ respectively. 
\subsubsection{Implementation}
To capture the article representation with respect to the introduced aspects above, we use the CP/PARAFAC and SVD to decompose the $\mathcal{X}_{w\times w\times n}$ and $\textbf{X}_{N\times P}$ into view matrices. In fact, in this case, the views are the factor matrices corresponding to the article mode and are of size $N\times r_m$ where $N$ and $r_m$ are the number of articles and the rank of decomposition respectively:
\begin{dmath}
\textbf{$\mathcal{X}_{TTA}$} \approx \Sigma_{r=1}^{R} \mathbf{a}_r \circ \mathbf{b}_r \circ \mathbf{c}_r
\end{dmath}
\begin{dmath}
X_{TAGS}\approx\mathbf{U} \boldsymbol{\Sigma} \mathbf{V}^{T}
\end{dmath}
\par After decomposing $\mathcal{X}_{TTA}$ and $\textbf{X}_{TAGS}$ into view matrices using CP/PARAFAC and SVD respectively, we apply the CCA on factor matrices $C$ and $U$ that represents the articles patterns. The result of CCA provides us with the canonical matrices where the row $i$ of these matrices correspond to datapoints $P_{1i}$ and $P_{2i}$ which could be leveraged to define a line or a 2-views representation of news article $i$. We construct a graph such that the lines are the nodes and the edges are defined as mean distances between the lines and the points on the other lines using the equation \ref{equ:point_line_distance}. Finally, to classify articles, we leveraged the belief propagation algorithm implemented in \cite{KoutraKKCPF11} to propagate 40\% of the labels throughout the KNH graph in a \textit{semi-supervised} manner.

\subsubsection{Experimental Result}
To evaluate the performance of proposed KNH in comparison to classic KNN, we create a KNN graph by calculating the Euclidean distance between the rows of each view matrix separately and model the similarity of articles by taking the average pairwise distances of points. In other words, to calculate the distance between articles $i$ and $j$, we calculate the pairwise Euclidean distance of rows $i$ and $j$ for both matrices $C$ and $U$ individually and then take the average of the resulted distances and consider it as the edge between node $i$ and $j$ in KNN graph. Moreover, to make the comparison between the KNH and KNN graphs fair enough, in another KNN model, we also project $C$ and $U$ using CCA into a maximally correlated space. Although this step is not required for KNN graph due to independency of views in KNN modeling, we apply CCA to minimize the effect of other pre-processes in classification performance. 
\par The average F1 score achieved by 10 runs of for different ranks of decomposition and number of neighbors $k$ are demonstrated in Figure \ref{fig:twitter}. The trend of F1 score for three  modelings, i.e., KNN, KNN after CCA and KNH graphs suggests that, KNH or manifold modeling of the articles using 2-flats (lines) in this case, leads to higher classification performance. As Shown, the highest performance achieved by rank 20 and for all three models. Thus, we report the precision, recall, F1-score and accuracy for this  $R$=20 and $k$=15 in Table. \ref{table:ourdata}.
\par As reported in Table. \ref{table:ourdata}, applying CCA before KNN modeling does not affect the results significantly due to independency of views in this approach. Moreover, the reported results of this table achieved by rank 20 where the KNN has the highest performance and the difference between the two models is minimum. However, this difference increases significantly when we increase the rank of decomposition which means when we capture more details of each view the manifold representation of KNH is more capable to take advantage of it.
\begin{table}[!th]
\footnotesize
\addtolength{\tabcolsep}{-4.5pt}
\begin{tabular}{
||p{2cm}p{1.5cm} p{1.5cm} p{1.5cm}p{1.5cm}||}
\hline
\centering
 \textbf{Method}&\textbf{Precision}&\textbf{Recall}&\textbf{F1 Score}&\textbf{Accuracy}\\
 \toprule
 \hline
 CP+KNN&0.687$\pm$0.008&0.683$\pm$0.011&0.684$\pm$0.008&0.694$\pm$0.007\\
 CP+CCA+KNN&0.691$\pm$0.007&0.682$\pm$0.018&0.686$\pm$0.011&0.697$\pm$0.011\\
 
 CP+CCA+KNH 
 &\textbf{0.709}$\pm$\textbf{0.011}&\textbf{0.720}$\pm$\textbf{0.016}&\textbf{0.713}$\pm$\textbf{0.012}&\textbf{0.719}$\pm$\textbf{0.011}\\
\bottomrule
\end{tabular}
\caption{Classification performance of KNH modeling against KNN modeling for $R$=20 and $K$=15 on Twitter dataset. The results suggest that regardless of whether we apply the correlation maximization on views or not the KNH outperforms the classic KNN.}
\label{table:ourdata}
\end{table}
\subsection{Experiment 2: Article Classification using 2-Flats (Lines) Created by User-News and Publisher-News Interactions Aspects}
\subsubsection{Description of Dataset and Aspects}
For the second experiment, we again aim at modeling the news articles but this time using aspects other than those of previous experiment and from a different dataset to examine the efficacy of KNH on aspects of different nature. To this end, we use the FakeNewsNet dataset\cite{shu2018fakenewsnet} \footnote{https://github.com/KaiDMML/FakeNewsNet} which consists of users and publisher information for news articles crawled from PolitiFact web site. The content of this website is typically shared on social media such as Twitter. The reason for using these two aspects is that in \cite{Beyond} the author have shown that these two aspects lead to promising result in terms of classification of articles. The details of the FakeNewsNet dataset is reported in table \ref{table:FakeNewsNet}.

\begin{figure*}[!th]
\centering
\subfigure[Average F1-score for 10 runs of decomposition and k=20 when modeling the articles by KNN and KNH graphs. As shown, KNH  modeling achieves higher F1 scores in comparison to KNN modeling.]{\includegraphics[width = 0.40\linewidth]{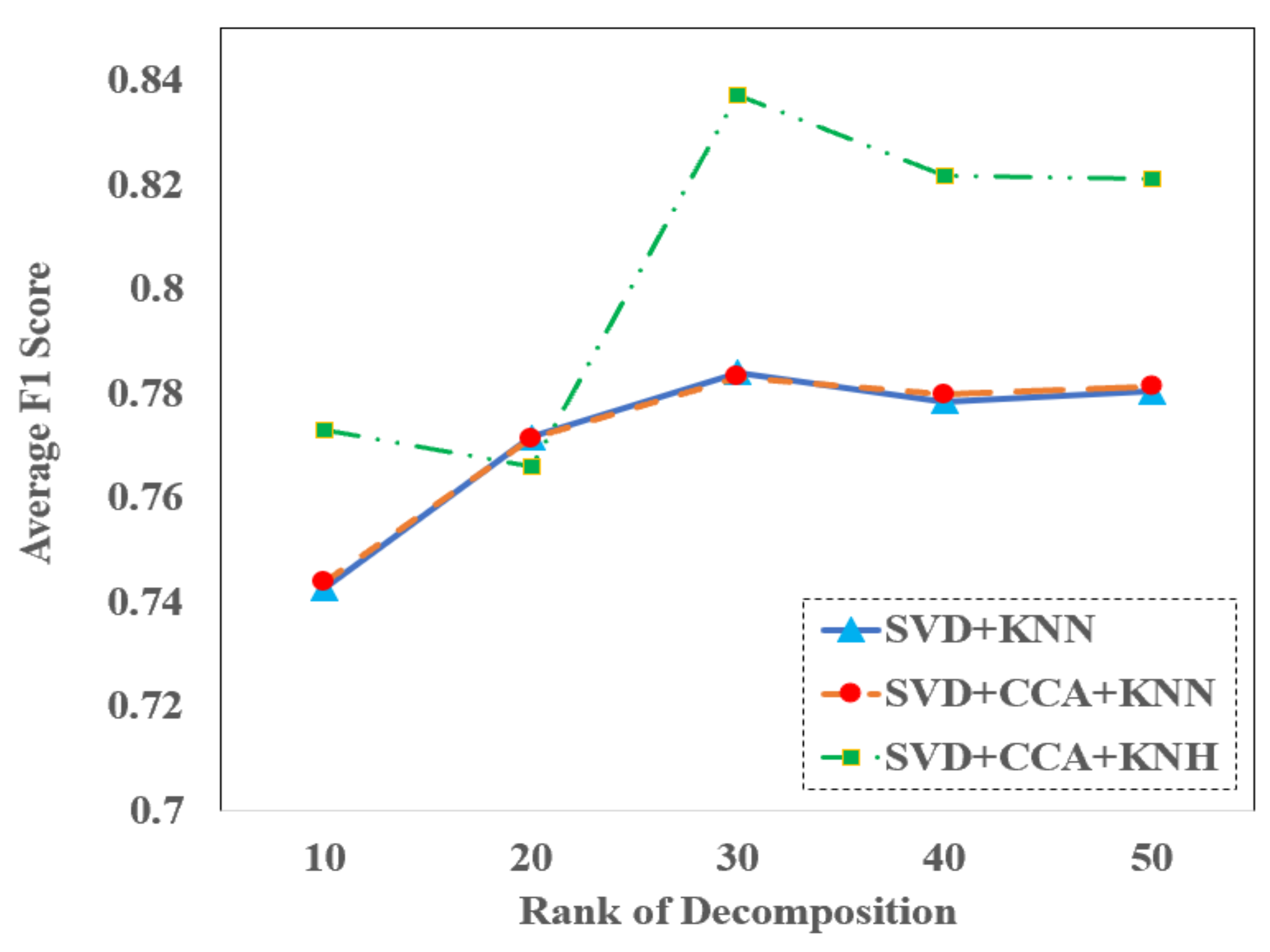}}
\hspace{15mm}
\subfigure[Average F1-score for 10 runs of decomposition and R=30 when modeling the articles by KNN and KNH graphs. As shown, KNH  modeling achieves higher F1 scores in comparison to KNN modeling.]{\includegraphics[width = 0.40\linewidth] {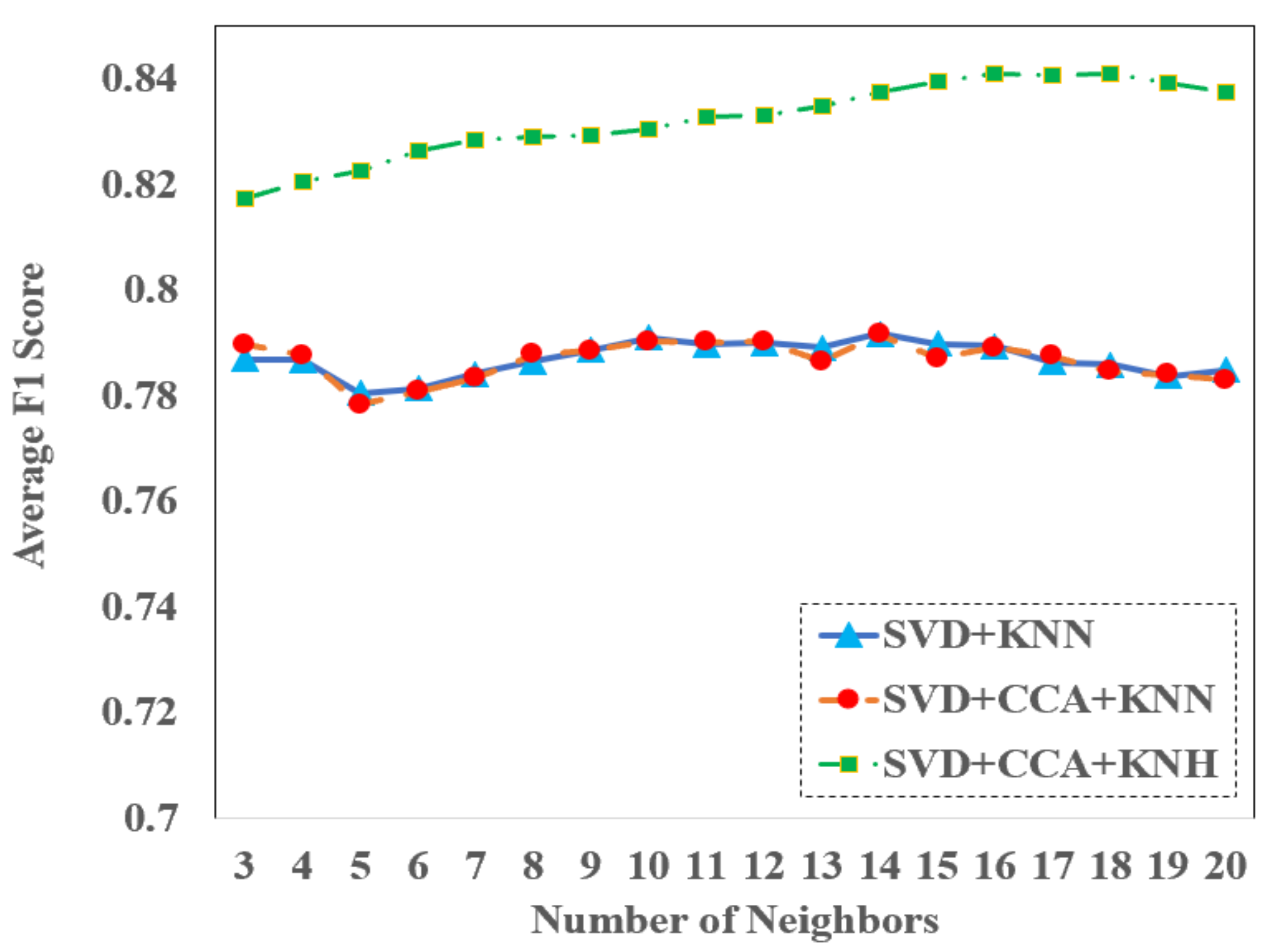}}
\caption{Average F1 score for different rank of decomposition $R$ and different number of neighbors $K$.}
\label{fig:politifact}
\end{figure*}
\begin{table}[!thb]
\centering
\footnotesize
\begin{tabular}{
||p{4.5cm}p{1.5cm} p{1.5cm}||}
 \hline
 \multicolumn{3}{||c||}{\textbf{FakeNewsNet dataset}} \\
 \toprule
 \hline
 \textbf{Features}&\textbf{Real}&\textbf{Fake}\\
 \hline
 Total news articles&432&624 \\
 Total number of tweets&116005&261262 \\ 
 Total news with social engagement&342&314\\
 Total number of Users&214049&700120 \\ 
\bottomrule
\end{tabular}
\caption{FakeNewsNet dataset description}
\label{table:dataset description}
\end{table}
\begin{table}[!th]
\addtolength{\tabcolsep}{-4.5pt}
\centering
\footnotesize
\begin{tabular}{
||p{2cm}p{1.5cm} p{1.5cm} p{1.5cm}p{1.5cm}||}
\hline
\centering
 \textbf{Method}&\textbf{Precision}&\textbf{Recall}&\textbf{F1 Score}&\textbf{Accuracy}\\
 \toprule
 \hline
 SVD+KNN
 &0.836$\pm$0.004&0.742$\pm$ 0.004&0.789$\pm$0.001&0.780$\pm$0.002\\
 SVD+CCA+KNN
 &0.833$\pm$0.003&0.747$\pm$0.002&0.787$\pm$0.001&0.777$\pm$0.001\\
 SVD+CCA+KNH&\textbf{0.875}$\pm$\textbf{0.002}&\textbf{0.808}$\pm$\textbf{0.002}&\textbf{0.839}$\pm$\textbf{0.001}&\textbf{0.830}$\pm$\textbf{0.001}\\ 
\bottomrule
\end{tabular}
\caption{Classification performance of KNH modeling against KNN modeling for $R$=30 and $K$=20 on FakeNewsNet dataset. The results suggest that regardless of whether we apply the correlation maximization on views or not the KNH outperforms the classic KNN.}
\label{table:FakeNewsNet}
\end{table}
For this experiment we use the following models as suggested in \cite{Beyond} to examine the efficacy of KNH  modeling in comparison to classic KNN:
\begin{itemize}
    \item \textbf{User-News Interaction}: As suggested in \cite{Beyond} We create a matrix to model the users who tweets a specific news article. The rows of this matrix are users and the columns are the news IDs.
    \item \textbf{Publisher-News Interaction}: We create a matrix to model the publishers that published a specific news article. The rows of this model are the publishers and the columns are the news IDs \cite{Beyond}.
\end{itemize}
\par Henceforth, we refer to the User-News interaction and the publisher-news interaction matrices as $X_{UN}$ and $X_{PN}$ respectively. 
\subsubsection{Implementation}
To capture the latent representation of articles in view spaces, we first decompose the $X_{UN}$ and $X_{PN}$ using SVD rank $r_m$ individually as follows:
\begin{dmath}
X_{PN}\approx\mathbf{U_1} \boldsymbol{\Sigma_1} \mathbf{V_1}^{T}   
\end{dmath}
\begin{dmath}
X_{UN}\approx\mathbf{U_2} \boldsymbol{\Sigma_2} \mathbf{V_2}^{T}
\end{dmath}

 Where $X_{UN}$ and $X_{PN}$ are of size $U\times N$ and $P \times N$ respectively and the $V_1$ and $V_2$  matrices are of size $N \times r_1$ and $N \times r_2$ and contain latent patterns of entities (news articles in this case). Then as explained earlier, we apply the CCA to transfer view matrices into a maximally correlated common space. Then we create a KNH graph in which the nodes are the lines or 2-flats in $r_m$ dimentional space and the edges are defined as the mean euclidean distance between the lines and the points lie on the other lines. Finally, just like the previous experiment, we leveraged the belief propagation algorithm to propagate 40\% of the ground truth in a semi-supervised manner.

\subsubsection{Experimental Result}
Again to compare the proposed KNH and the classic KNN graph, we follow the same strategy to calculate the similarity of article $i$ and $j$. In other words, we calculate the Euclidean distance of rows $i$ and $j$ for both matrices $V_1$ and $V_2$ and then take the average of the resulted distances. Likewise the previous experiment, to have a fair comparison between the KNH and KNN graphs we also report the results of KNN after projection using CCA. The average F1 score achieved by 10 runs of these experiments for different ranks of decomposition are demonstrated in Figure \ref{fig:politifact}. 
\par This experiment also yields to similar results i.e., The trend of F1 score for the three different modelings, suggests that, manifold modeling of the articles using 2-flats or lines, leads to higher classification performance. As illustrated, best results achieved by rank 30 for all models. Classification metrics for $R$=30 and $K$=20 are reported in Table. \ref{table:FakeNewsNet}. Like previous experiment, by increasing the rank, KNH modeling achieves higher performance than KNN graphs which again suggest that KNH is more capable of consolidating details of views.
\section{Conclusion and Future Work}
In this work, we introduce a novel multi-view modeling of the entities (articles) by generalizing the classic KNN graph. We propose to model nodes of the graph as hyperplanes (m-flats) using datapoints derived from different views of the articles and then suggest a way to define the edges between hyperplanes. We experiment the proposed K-Nearest Hyperplane graph (KNH) on two different 2-aspect datasets. The experimental results suggest that for different ranks and number of neighbors KNH graph outperforms the classic KNN graph. 
However, there are many possible directions for improving the idea of this work. Some of them are as follows:
\begin{itemize}
\item As discussed in background section, we can leverage parametric equations of hyperplanes for formulating and representing the m-view entities by m-flats. We experimented on 2 different 2-aspect datasets. However, by increasing the number of views we require more mathematical tools  to calculate requirements of the Euclidean subspaces e.g. cross product in higher dimensional space. Unfortunately, due to the space limitation we are not able to discuss it in details. We reserve the higher view formulation of this work for future work. Moreover, as mentioned earlier, a rationale behind defining a common space for multi-view nodes in addition to a consolidate representation of the enities is to take advantage of this common space for estimating missing or unknown features that may fall into this common space. In future work, we will also explore the capability of KNH graph for prediction of missing features.
\item Even though we defined the simplest way for defining the edges or multi-view similarity of nodes in KNH graphs in this work, we are interested in defining more insightful edges between the nodes by probably merging the capabilities of hypergraphs that take into account higher order relations between the nodes and the advantages of multi-aspect nodes of this work. We reserve the study and formulation of more meaningful edges for future work. 
\end{itemize}
\section{ACKNOWLEDGEMENTS}
Research was supported by a UCR Regents Faculty Fellowship, a gift from Snap Inc., the Department of the Navy, Naval Engineering Education Consortium under award no. N00174-17-1-0005, and the National Science Foundation Grant no. 1901379. The authors would like to thank Rutuja Gurav for her invaluable help with the proofreading of the paper. Any opinions, findings, and conclusions or recommendations expressed in this material are those of the author(s) and do notnecessarily reflect the views of the funding parties.
\balance

\end{document}